\documentclass[final,5p,times]{elsarticle}
\usepackage{amssymb}
\usepackage{graphicx}
\usepackage{lineno}
\usepackage{caption,subcaption}
\usepackage{multirow}
\usepackage{caption}
\usepackage{subcaption}
\usepackage{array}
\usepackage{threeparttable}
\usepackage{amsmath}
\usepackage{nccmath} % equation
\usepackage{xcolor}
\usepackage{hyperref}
\hypersetup{colorlinks=true,
linkcolor=black,
anchorcolor=red,
citecolor=cyan,
filecolor=red,
urlcolor=red,
pdfauthor=author
}
\journal{Applied Energy}

\begin{document}

\begin{frontmatter}



\title{A lightweight network for photovoltaic cell defect detection in electroluminescence images based on neural architecture search and knowledge distillation}
% based on neural architecture search and knowledge distillation
% \author{}
% \fnref{fn1}}
\author[seu,address2]{Jinxia Zhang\corref{cor1}}
\ead{jinxiazhang@seu.edu.cn}
\cortext[cor1]{Corresponding author}
\author[seu]{Xinyi Chen}

\author[seu]{Haikun Wei}
\author[seu]{Kanjian Zhang}
% affiliation address
%\address[seu]{organization={Key Laboratory of Measurement and Control of CSE, Ministry of Education, School of Automation},
%             addressline={Southeast University}, 
%             city={Nanjing},
%             postcode={210096}, 
%             state={Jiangsu},
%             country={China}}
% \address[address2]{organization={Southeast University Shenzhen Research Institute},
%             city={Shenzhen},
%             postcode={518057}, 
%             state={Guangdong},
%             country={China}}
\address[seu]{Key Laboratory of Measurement and Control of CSE, Ministry of Education, School of Automation, Southeast University, Nanjing, 210096, Jiangsu, China}
\address[address2]{Southeast University Shenzhen Research Institute, Shenzhen, 518057, Guangdong, China}
\begin{abstract}
Nowadays, the rapid development of photovoltaic(PV) power stations requires increasingly reliable maintenance and fault diagnosis of PV modules in the field. Due to the effectiveness, convolutional neural network (CNN) has been widely used in the existing automatic defect detection of PV cells. However, the parameters of these CNN-based models are very large, which require stringent hardware resources and it is difficult to be applied in actual industrial projects. To solve these problems, we propose a novel lightweight high-performance model for automatic defect detection of PV cells in electroluminescence(EL) images based on neural architecture search and knowledge distillation. To auto-design an effective lightweight model, we introduce neural architecture search to the field of PV cell defect classification for the first time. Since the defect can be any size, we design a proper search structure of network to better exploit the multi-scale characteristic. To improve the overall performance of the searched lightweight model, we further transfer the knowledge learned by the existing pre-trained large-scale model based on knowledge distillation. Different kinds of knowledge are exploited and transferred, including attention information, feature information, logit information and task-oriented information. Experiments have demonstrated that the proposed model achieves the state-of-the-art performance on the public PV cell dataset of EL images under online data augmentation with accuracy of 91.74\% and the parameters of 1.85M. The proposed lightweight high-performance model can be easily deployed to the end devices of the actual industrial projects and retain the accuracy.
\end{abstract}
%Graphical abstract
% \includegraphics[scale=0.6]{fig/method.pdf}

% \begin{highlights}
% \item A novel state-of-the-art lightweight model is proposed for automatic defect detection in electroluminescence images. 
% \item The lightweight model structure is automatically designed by neural architecture search to save manual workload.
% \item The multi-scale characteristic of photovoltaic cell
% defects is taken into consider for search space design.
% \item To make full use of priors learned by large network, different knowledge is transferred into the searched lightweight network.
% \item Extensive experiments on public dataset and private dataset demonstrate the effectiveness of the proposed method and the potential for practical terminal deployment.
% \end{highlights}
\end{frontmatter}
\section{Introduction}
The lifetime of photovoltaic(PV) modules is essential for power supply and sustainable development of solar technology. However, the PV cells are easily affected by various external factors. During the manufacturing process, minor operational errors may result in module damages. In addition, vibration and shock during transportation and installation may also cause module breakage. Defects such as cracks, solder corrosion, cell interconnect breakage can make PV modules unusable, and microcracks that are hard to observe will potentially affect future output power and lifetime\cite{NDIAYE2013140,li2019thermo}. The above defects in PV cells may cause module failure during operation, which can lead to power reduction and even safety problems for the whole system\cite{kontges2014review}.
\par
The current-voltage(I-V) curve is used for detection of defective PV modules. Changes in I-V characteristics can reflect those heavily degraded modules. However, tiny cracks can hardly affect I-V characteristics, thus are difficult to identify. These microcracks have the potential possibility of separation and degradation, which can seriously affect the future use\cite{kajari2012criticality}.
As described in some research, microcracks can cause power attenuation, the loss of which varies from 0.9\% to 42.8\%, and may cause hot spot effect\cite{dhimish2020micro,abdelhamid2013review}.
Besides I-V curve, infrared thermal(IRT) imaging\cite{TSANAKAS2016695} is another technology which can be used to detect defects.
The temperature of PV cells with defects is significantly higher than other cells around them. However, the hot spots of the PV modules are not necessarily caused by the defects. Other factors like object occlusion can also lead to the abnormal detection results. Also, microcracks which have not yet affecting power efficiency can not be recognized by IRT images with a relatively low resolution. 
\par
Due to the high resolution of imaging, electroluminescence (EL) imaging\cite{fuyuki2009photographic} has become one of the most commonly used methods for defect detection of PV modules.
EL imaging system is a non-destructive technology with high imaging resolution which can be used to detect microcracks\cite{breitenstein2011can}.
In EL images, cracks and other defects in defective PV cells appear as dark gray lines and areas. 
In early stage, traditional methods based on manual features are proposed to detect defects in EL images. These methods depend on large amounts of manual design experiments and the performances are limited.
\par
Because of the strong feature capturing ability of convolutional neural network(CNN), the methods using deep learning have gradually become the mainstream to detect defects in EL images. However, while CNN has greatly improved the detection accuracy, it also requires more time and hardware resources, making it hard to be deployed in end devices of practical applications. In order to meet the requirements of both the
accuracy and speed of defect detection in the industrial field, a lightweight and efficient detection network is required.\par
There are a few works\cite{karimi2019automated,tang2020deep,akram2019cnn} which have proposed \\lightweight CNN-based methods to detect defective PV cells in EL images. These lightweight CNN-based methods are all based on manual design, which require a lot of experiments to find a suitable network structure. 
To obtain a lightweight structure for practical application with much less manual work, we introduce neural architecture search (NAS) into the defective PV cell classification task, which is the first method using NAS to automatically design networks in the field of PV defect detection.
Aiming at the automatic design of network architecture, NAS can reduce manual intervention and make better use of computing resources in an automated manner.
Since the defects can be any size, we propose a search space which can enhance features at different scales, obtaining a lightweight network architecture that can better extract multi-scale features.
\par
To make better use of the prior knowledge, knowledge distillation is introduced to learn the priors obtained by the existing pre-trained large-scale model to improve the performance of the searched lightweight network. Different kinds of knowledge are transferred, including attention information, feature information, logit information and task-oriented information. The obtained lightweight network has a high performance, which even outperforms the existing large-scale teacher model.
\par
The contributions of the proposed method can be summarized as follows:
\begin{enumerate}[1)]
\item We propose a lightweight network structure for detection of defective PV cells with high accuracy of 91.74\% and size of 1.85M parameters, achieving the state-of-the-art performance on public PV cell dataset\cite{ELPV} of EL images under online data augmentation. The proposed model also has high accuracy on defective PV cells up to 94.26\% on our private dataset.
\item We introduce NAS to the field of PV cell defect detection for automatic lightweight network design, which reduces the workload of manual design. To detect defects with any size, the search space is designed by considering multi-scale characteristic into the network architecture.
\item To make full use of the priors already learned by the existing large-scale network, we utilize knowledge distillation to transfer various prior knowledge into our model. We consider attention information, feature information, logit information and task-oriented information into the knowledge transfer process and the experiments prove the effectiveness of knowledge distillation to enhance the ability of recognizing defective PV cells.
\end{enumerate}
\section{Related work}
\subsection{Traditional methods}
Some research uses traditional image processing methods to detect defects in EL images. These methods usually rely on the manually selected features.
\par
Dhimish et al.\cite{dhimish2019novel,dhimish2019solar} used the bit-by-bit OR gate method to process EL images and enhance crack images, but the detection accuracy and other results were not given. Tsai et al.\cite{tsai2012defect} presented an independent component analysis technique, but finger cracks that have little effect on crack detection were identified as other cracks. Anwar et al. \cite{anwar2014micro} proposed an improved image segmentation method based on anisotropic diffusion filter and support vector machine(SVM) was employed to detect micro-crack defects based on medium-sized datasets, but this method requires a higher level of pre-processing. Su et al.\cite{su2019classification} improved a new feature descriptor, which combines central pixel gradient information with central symmetric local binary mode to obtain more recognizable defect features under uneven background interference. 
\par
In traditional image processing methods, edge gradient information is often used to describe the features. Due to the similarity between the change of edge gradient of defects and the grain under complex background, it is easy to be disturbed when distinguishing defects from background grains. At the same time, these methods tend to be applied only on small datasets, and their generalization ability is not strong. 
\subsection{Deep learning based methods}
Due to the popularity of deep learning, surface defect detection of PV cells based on deep learning has become a research hotspot in this field. CNN is becoming a widely used detection method because of its strong feature extraction ability. 
\par

Sun et al.\cite{sun2017defect} proposed a crack classification network based on LeNet5\cite{LeNet5}, which can classify four kinds of crack defects. Bartler et al.\cite{bartler2018automated} 
designed an improved classification network based on VGG16\cite{vgg} structure and explored the effects of a few oversampling and data expansion methods on performance improvement. Deitsch et al.\cite{deitsch2019automatic} conducted two defect classification methods based on VGG19 and SVM, and contributed a PV cell dataset of EL images. Shou et al.\cite{shou2020defect} presented an unsupervised defect detection method based on generative adversarial networks(GAN), but the stability of the network needs to be further discussed. Both studies by Liu et al.\cite{liu2019surface} and Su et al.\cite{su2020deep} improved the 
%RPN
region proposal network on the basis of Faster-RCNN, and realized the detection of small cracks in PV cells.
\par
These studies are based on existing networks by transfer learning or improvement on some layers and parameters. Compared with traditional methods using EL images, deep learning methods have better generalization ability and higher accuracy.
%-------------
\subsection{Lightweight methods}
Most of the existing deep learning models are large, requiring high hardware deployment in field for PV cell defect detection.
\par
To solve this problem, a few researchers proposed lightweight networks by manual design. Karimi et al.\cite{karimi2019automated} designed a 4-layer CNN structure for classification of 3 kinds of defects. Tang et al.\cite{tang2020deep} designed a 9-layer CNN structure and improved the performance with a mixture of GAN generation and traditional data augmentation. Inspired by VGG11, a 9-layer CNN structure was designed by Akram et al.\cite{akram2019cnn} and validated on the public PV cell dataset\cite{ELPV}. 
Wang et al.\cite{WANG2022119203} 
utilized octave convolution to build a lightweight network with high inference speed. All these studies are based on manual network structure design, which are difficult and require a large amount of experiments. Besides, manual structure design depends a lot on the existing data and is less universal. 
\vspace{1em}
%-------
\par
To reduce the manual workload in model design, we introduce neural architecture search(NAS) into the task of PV cell classification for effective automatic architecture design. For better model training, we transfer different prior knowledge already learned by large-scale model based on knowledge distillation. In this process, attention information, feature information, logit information and task-oriented information are exploited and transferred to enhance the performance of the searched lightweight model.
\section{Methodology}
\begin{figure}[ht]
\centering
\includegraphics[scale=0.29]{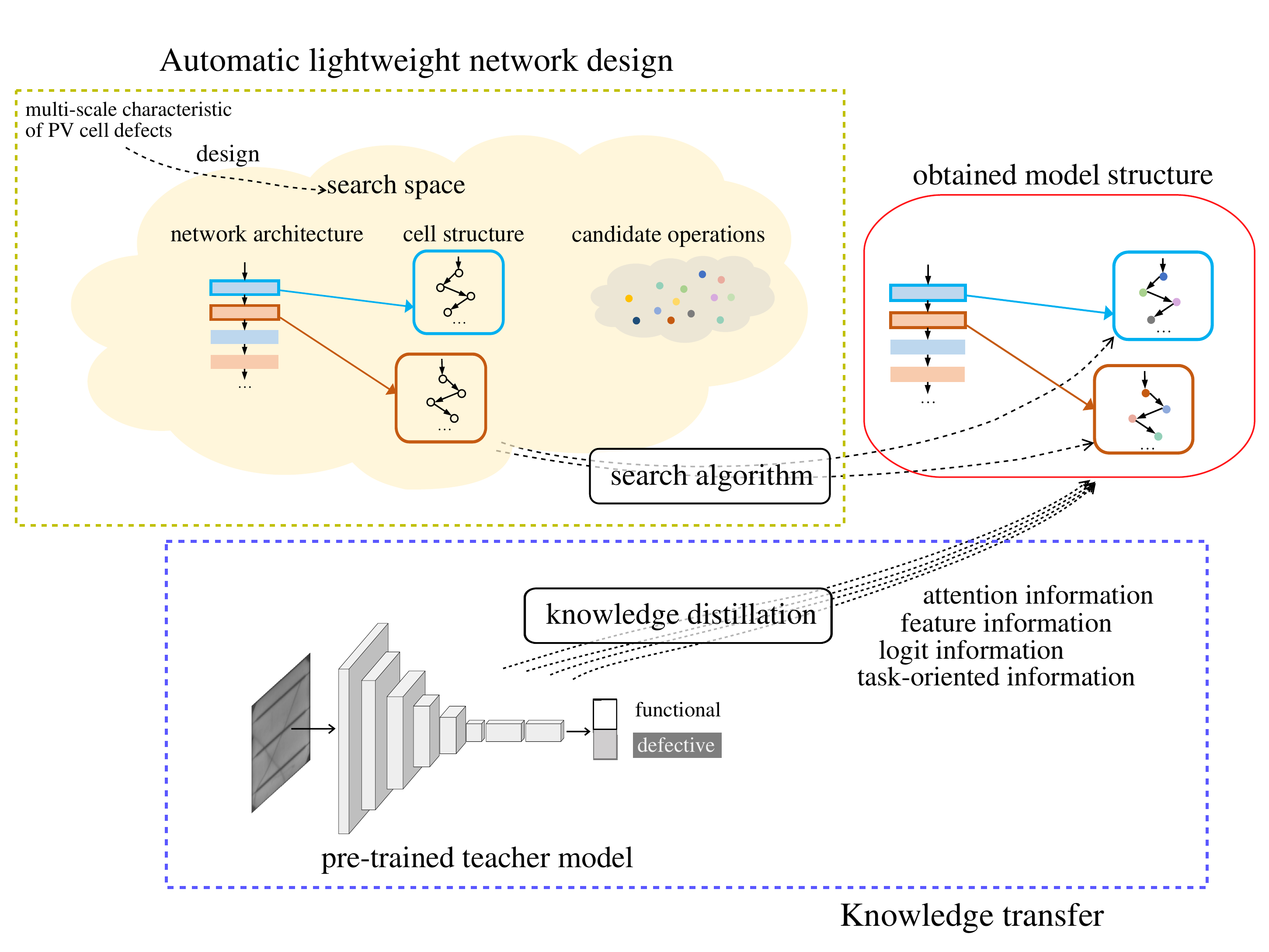}
\caption{The architecture of the proposed lightweight network design. Firstly, the lightweight network is automatically obtained by NAS algorithm in a designed search space. Then, the priors learned by the large network are transferred to the lightweight network through knowledge distillation.}
\label{overview}
\end{figure}

An effective lightweight network is proposed in this section for detection of defective PV cell by NAS and knowledge transfer. 
% To automatically design lightweight network, NAS is introduced to the filed of PV cell defect detection for the first time. 
To automatically design lightweight network, NAS is introduced to the field of PV cell defect detection for the first time.
To detect defects with any size, the network architecture search space is designed by adding multi-scale characteristic. Then a variety of prior knowledge is transferred by knowledge distillation to make full use of the priors already learned by the large-scale network. The illustration of our method is depicted in Figure~\ref{overview}.
\subsection{Automatic lightweight network design}
We employ a continuous gradient-based NAS framework, i.e. DARTS\cite{liu2018darts} to design the lightweight network automatically for PV cell defect detection
since DARTS has a fast search speed. 
We further design a suitable search space by consideration of the visual
multi-scale characteristic of the PV cell defects.
% We further design a suitable search space by taking consider of the visual multi-scale characteristic of the PV cell defects.
\par
The defects of PV cells can be of any size, and the tiny microcrack detection in particular is a difficult issue. 
Considering the multi-scale characteristic of the defects, the search space is designed to enhance features of different size. 
\par
The employed search space for the lightweight network is mainly stacked by two kinds of cell structures called normal cell and reduction cell% The empolyed search space for the lightweight network is mainly stacked by two kinds of cell structures called normal cell and reduction cell
, which is based on the idea of reusing blocks like ResNet\cite{resnet}. The possible connection type of nodes in cell structure will be chosen from candidates operations. 
\par
\begin{figure}[ht]

\centering
\includegraphics[scale=0.8,angle=90]{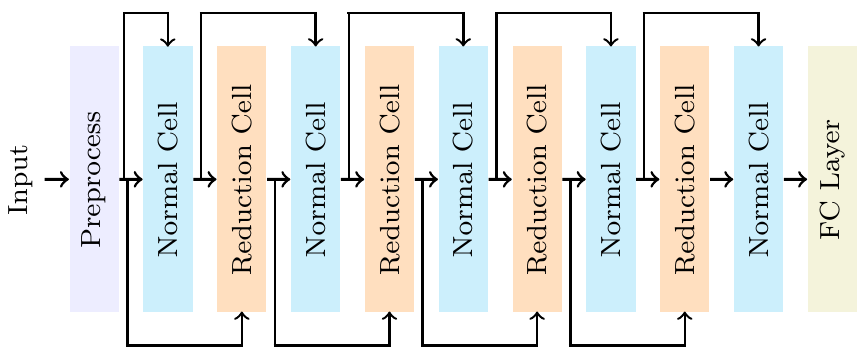}
\caption{Designed search space for the lightweight defect classification network.}
\label{mymodel}
\end{figure}

Cell structure stacked in network architecture can be considered as a kind of convolutional operations. Normal cell is set to maintain the size of the input, while reduction cell has the function of down-sampling. To obtain multi-scale information, the search space for the lightweight network architecture is designed by stacking five normal cell and four reduction cells. The designed search space for the PV cell defect recognition task is shown in Figure~\ref{mymodel}.  
Each cell fuses two features with different scales of the previous two cells, and the first normal cell takes the same feature twice as two inputs.  

\par
The details of the lightweight network is presented in Table~\ref{netstructure}. Here the first three reduction cells perform downsampling and channel expansion, while the channel number of the last reduction cell remains the same. The proposed network will finally classify the input PV cell as functional or defective. 
\begin{table}[ht]
\centering
\setlength{\tabcolsep}{2mm}{
\scriptsize
\renewcommand\arraystretch{1.4} 
\begin{tabular}{l l}
\hline
Layer     &  Output shape     \\ \hline
Input&3$\times$150$\times$150\\
Preprocess(3$\times$3 Convolutional Layer with stride 1)&64$\times$150$\times$150  \\
Normal Cell& 64$\times$150$\times$150   \\
Reduction Cell& 128$\times$75$\times$75  \\
Normal Cell& 128$\times$75$\times$75  \\
Reduction Cell&  256$\times$38$\times$38 \\
Normal Cell& 256$\times$38$\times$38  \\
Reduction Cell&  512$\times$19$\times$19 \\
Normal Cell& 512$\times$19$\times$19 \\
Reduction Cell&  512$\times$10$\times$10 \\
Normal Cell&512$\times$10$\times$10   \\
Global Average Pooling(GAP)&512$\times$1$\times$1  \\
Fully Connected(FC) Layer&2 \\
\hline
\end{tabular}
}
\caption{The detailed structure of the proposed lightweight network.}
\label{netstructure}
\end{table}
%-------------------------------------------

\begin{figure}[ht]
\centering
%cellmm
\includegraphics[scale=1]{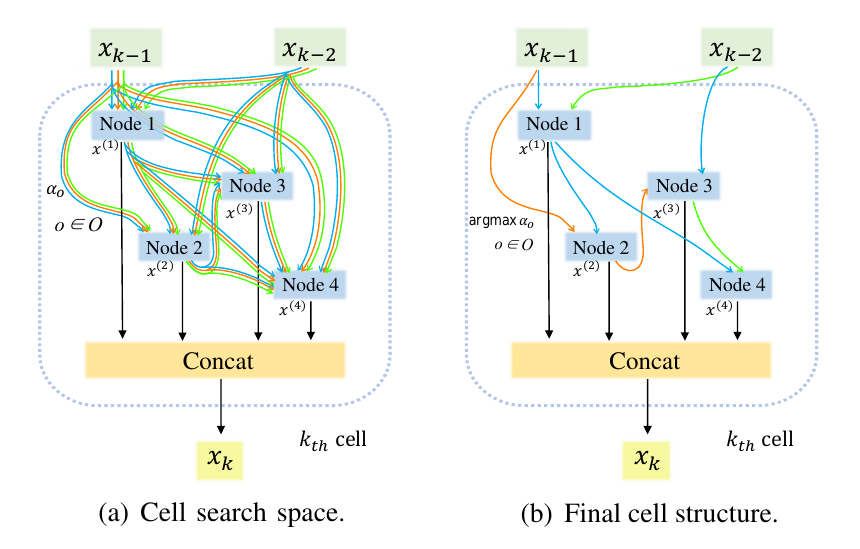}
\caption{Internal structure of cells. It is an example of the $k_{th}$ cell with 4 internal nodes. Colored lines between two nodes represent the different candidate operations. (a) shows the search space in cell structure. Each node gets a mixture of features by different candidate operations. (b) is the final structure determined by the structure weight computed in the search algorithm.}
\label{cellin}
\end{figure}
\par
The internal structure of the cell is a directed acyclic graph containing $N$ nodes, where each node represents the computed temporary feature map as in Eq.~\eqref{eq1}. Let $x_{}^{(j)}$ denote the computed temporary feature at the $j_{th}$ node. 
Each node is computed based on the input feature maps from two previous nodes.
\begin{center}
\begin{equation}
\label{eq1}
x^{(j)} =\sum_{i<j}^{} o^{(i,j)}(x^{(i)})
\end{equation}
\end{center}
\par
Figure~\ref{cellin} shows the example of the $k_{th}$ cell with 4 internal nodes. Different operations ${o}^{(i,j)}$ denoted as colored lines represent the function of candidate operations between the $i_{th}$ node and the $j_{th}$ node as shown in Figure~\ref{cellin}.(a). 
\par
$\mathcal{O}$ is a set of candidate operations which are listed in Table~\ref{operations}. The candidate operations can be mainly summarized into convolutional and other kinds. Convolutional layers of the candidate operations consist of depthwise separable convolution(SepConv) and dilated depthwise separable convolution(DilConv) with two optional values of the stride. In normal cells, the stride of convolutional layers is set as 1 to remain the size of the input. But in reduction cells, the stride is set as 2 for down sampling.
The setting of stride varies in two kinds of cells, which results in the different output feature size.

\par
The feature graph computed by all operations from the $i_{th}$ node to the $j_{th}$ node is calculated as in Eq.~\eqref{eq2}. The search space become continuous with the softmax transformation of each candidate operation between pairs of nodes $(i,j)$, where the weight of candidate operation $o_{}^{(i,j)}$ is denoted as $\alpha_{o^{} }^{(i,j)}$. 
% 两个输入
\begin{center}
\begin{equation}
\label{eq2}
\bar{o}^{(i,j)}(x) =\sum_{o\in \mathcal{O} }\frac{exp(\alpha_{o}^{(i,j)} )}{ {\textstyle \sum_{o^{'}\in \mathcal{O} }^{} exp(\alpha_{o^{'} }^{(i,j)})}  }o(x) 
\end{equation}
\end{center}
\par
The search framework of DARTS focuses on the learning of a set of structure variables $\alpha = \{\alpha_{normal}^{},\alpha_{reduction}^{}\}$ which denotes the weight of each candidate operation. The choice of final structure is obtained as in Eq.~\eqref{eq3}. 
\begin{center}
\begin{equation}
\label{eq3}
o^{(i,j)} =argmax\,\alpha_{o}^{(i,j)},o\in \mathcal{O}
\end{equation}
\end{center}
\par 
The operations with the highest structure weight will be the final choice in the network. As shown in Figure~\ref{cellin}.(b), the two operations of each node with highest structure weight are selected as the final cell structure.
\par
The learning process of $\alpha$ is treated as a bilevel optimization problem, with structure $\alpha$ as the upper-level variable and weights
$\omega$(weights of the convolution filters) as the lower-level variable. Because of the continuity of the search space, the gradient-based strategy optimizes the final structure $\alpha^*$ by minimizing the validation loss and optimizes the weight $\omega$ by minimizing the training loss simultaneously\cite{liu2018darts}:
\begin{center}
\begin{equation}
\label{eq4}
\begin{array}{ll}
\min _{\alpha} & \mathcal{L}_{\text {val }}\left(\omega^{*}(\alpha), \alpha\right) \\
\text { s.t. } & \omega^{*}(\alpha)=\operatorname{argmin}_{w} \mathcal{L}_{\text {train }}(\omega, \alpha)
\end{array}
\end{equation}
\end{center}
\par
%----------------operations
\begin{table}[t]
\centering
\begin{threeparttable}
\setlength{\tabcolsep}{6mm}{ 
\scriptsize
\renewcommand\arraystretch{1.4} 
\begin{tabular}{l l}
\hline
Type & Parameters     \\ \hline
\multirow{4}{*}{Convolutional}&SepConv\tnote{1}\quad(kernel=3,stride\tnote{2},padding=1)\\
&SepConv\quad(kernel=5,stride,padding=2)\\
&DilConv\tnote{3}\quad(kernel=3,stride,padding=2,dilation=2)\\
&DilConv\quad(kernel=3,stride,padding=2,dilation=2)\\
\hline
\multirow{3}{*}{Other}&Max Pooling\\
&Average Pooling\\
&Skip-connect\\
\hline
\end{tabular}
}
 \begin{tablenotes}
        \footnotesize
        \item[1]SepConv represents depthwise separable convolution.
        \item[2]Stride is set as 1 in normal cells and as 2 in reduction cells.
        \item[3]DilConv represents dilated depthwise separable convolution.
        
      \end{tablenotes}
  \end{threeparttable}

\caption{Candidate operations for structure search space.}
\label{operations}
\end{table}
%----------------------------------------------------
\subsection{Knowledge transfer}
\begin{figure*}[t]
\centering
\includegraphics[scale=0.8]{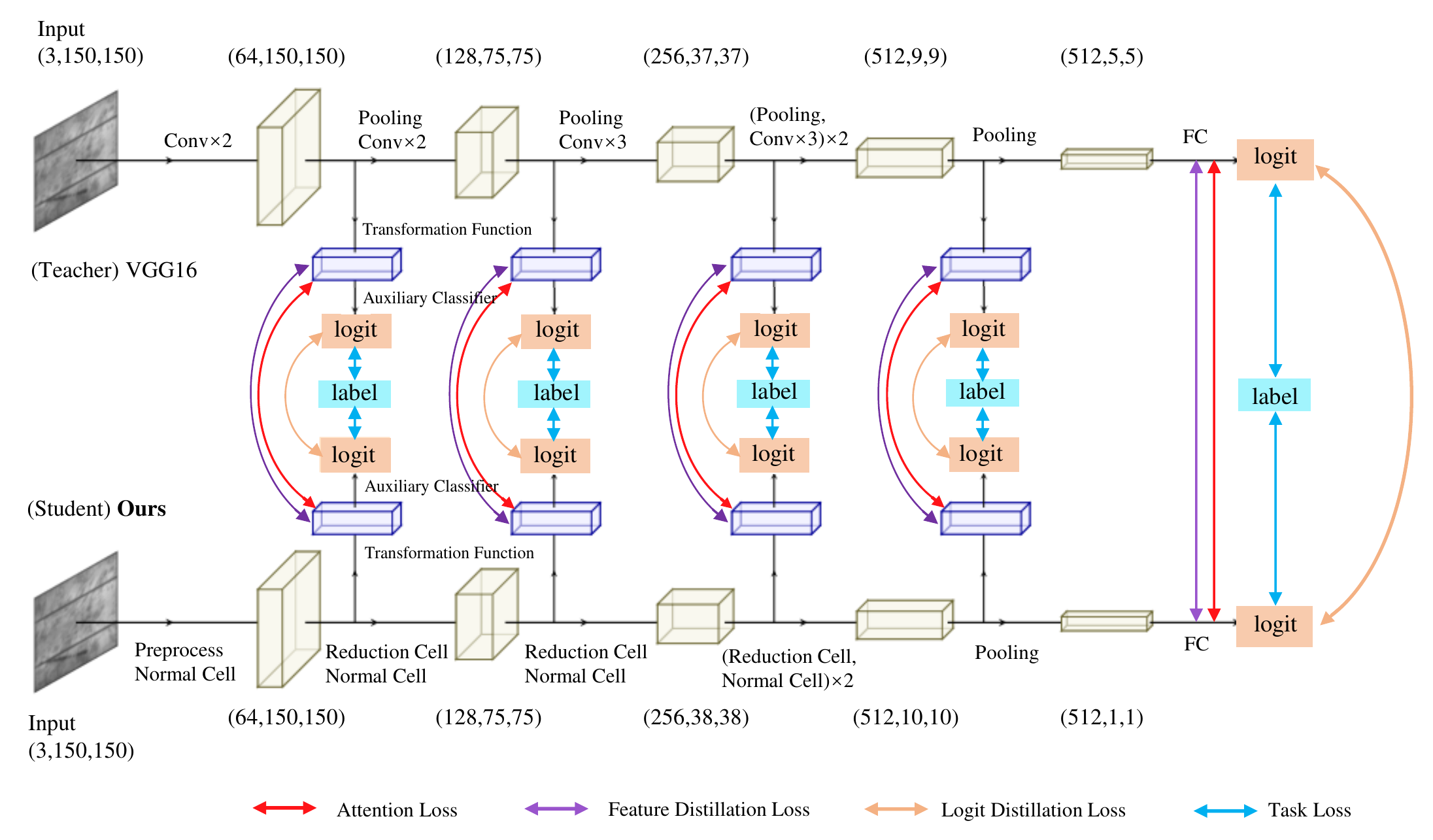}
\caption{The overview of the knowledge transfer. Teacher model and our student model are attached with auxiliary classifiers at each feature of different size. The diagram of feature transformation are clearly revealed. Features at different depth are selected for transfer by using auxiliary classifiers. The colored arrows point out the different loss components in the whole transfer process, including attention information, feature information, logit information and task-oriented information. The operations in transformation functions and auxiliary classifiers are only activated in knowledge transfer to capture target information.}
\label{kd}
\end{figure*}
%=---------
To make full use of the priors already learned by existing large-scale network and further improve the performance of the lightweight network architecture obtained by search process, several kinds of knowledge priors are exploited and transferred from the large network to the lightweight network.
\par
Knowledge distillation is one of the most effective methods
for model compression. It enables the
transfer of knowledge from a teacher model to a student model. Networks that cannot use the the prior knowledge in the pre-trained model can improve performance by learning the knowledge of the teacher network. Since the lightweight network can only be trained from scratch,
by using knowledge distillation, the priors can be utilized for better training.
\par
Inspired by different knowledge distillation works\cite{hinton2015distilling,gotmare2018closer,romero2014fitnets,zhang2020TOFD,zagoruyko2016AT}, four different knowledge priors are transferred: attention information, feature information, logit information and task-oriented information
in order to enhance the distillation effect of the PV cell defect detection task. 
\par  
The details of the knowledge distillation process are shown in Figure~\ref{kd}. 
 Transformation functions are used to exact useful information. The shape of features in teacher model and student model are usually different, and thus transformation can also make pairs of features to match the shape in computation. Attention map of each feature is introduced in transfer process. By comparing the distance of transformed features, the feature information are also utilized. Output logit provides information of logit prediction, which is also added into knowledge transfer. Convolutional layers and pooling layers are chosen as transformation functions to capture the task-oriented information from original feature map. 
The knowledge transfer paths are established before the feature map channel expanding. Knowledge transfer is carried out through the auxiliary classifiers at feature maps of different resolution, ensuring knowledge learning containing both low-level information and the high-level information. Note that the auxiliary classifiers are used only in the distillation process, not affecting the inference stage.
\par
Let $x_{i}$ represent the $i_{th}$ input of total $m$ images. Total $N$(in our method $N=5$) feature maps are selected into the transfer process. The $j_{th}$ feature map with different resolution is denoted as $F_{j}(\cdot)$. 
\par
%------------attention map
  Attention information is provided by the spatial attention map. To get a spatial attention map from the corresponding feature, the mapping function $A_{}(\cdot)$ through the channel dimension is applied, as illustrated in Eq.~\eqref{a}. 
  \begin{center}
\begin{equation}
\label{a}
A^{}(x)=\frac{1}{K} \sum_{k=1}^{K}||x_{k}{||}^{p}
\end{equation}
\end{center}
\par
In this way, the attention map can reflect the neuron activation spatially. 
  When the feature $x$ has $K$ channels, the attention map $A_{}(x)$ consists of average of absolute values of the feature 
  map across channel dimension, and each value is raised to the power of $p$. The operations of it are elementwise.
\par
The sum and power operation makes attention map focus more on spatial locations with high activations, i.e., the more discriminative parts. The attention loss $\mathcal{L}_{attention}$ is denoted in Eq.~\eqref{lat}. 
\begin{center}
\begin{equation}
\label{lat}
\mathcal{L}_{attention}=\frac{1}{m}\sum_{i=1}^{m}\sum_{j=1}^{N} L_{MSE}\left(A_{}^{}\left(T_{j}\left(F_{j}^{t}\left(x_{i}\right)\right)\right), A_{}^{}\left(T_{j}\left(F_{j}^{s}\left(x_{i}\right)\right)\right)\right)
% A^{}(x)=\frac{1}{K} \sum_{k=1}^{K}\left||x_{k}\right||^{p}
% \end{array}
\end{equation}
\end{center}
\par
Mean square error(MSE) is used to compute the distance between the attention maps of teacher model and student model as shown in Eq.~\eqref{lat}. Here the transformation function used to extract features is recorded as $T_{j}(\cdot)$.
% and $c_{j}(\cdot)$ respectly. 
The superscript $s$ and $t$ are used in order to distinguish the teacher model and the student model.
\par
Besides attention, feature maps also contain important information for knowledge distillation to improve the performance. 
\begin{center}
\begin{equation}
\label{l2}
\mathcal{L}_{\text {feature }}=\frac{1}{m}\sum_{i=1}^{m} \sum_{j=1}^{N} L_{2-norm}\left(T_{j}\left(F_{j}^{t}\left(x_{i}\right)\right), T_{j}\left(F_{j}^{s}\left(x_{i}\right)\right)\right)
\end{equation}
\end{center}
\par
The feature distillation loss uses L2-norm loss($L_{2-norm}^{}$) to compute the distance between each pair of features from teacher and student as shown in Eq.~\eqref{l2}. 
%-------------------------------logit
\par
Inspired by task-oriented feature distillation\cite{zhang2020TOFD}, we extract information for classification tasks by building auxiliary classifiers between features at different depths. The task loss function is presented in Eq.~\eqref{taskloss}.
\begin{center}
\begin{equation}
\label{taskloss}
\mathcal{L}_{\text {task }}=\frac{1}{m}\sum_{i=1}^{m} \sum_{j=1}^{N}\left.L_{C E}\left(c_{j}\left(T_{j}\left(F_{j}^{s}\left(x_{i}\right)\right)\right), y_{i}\right)\right)
\end{equation}
\end{center}
\par
In Eq.~\eqref{taskloss}, the $j_{th}$ auxiliary classifier is recorded as $c_{j}(\cdot)$. It returns the classification result as a vector.
%---------------------------task loss
% task loss
Cross entropy loss($L_{C E}$)is used here to compute the prediction logits obtained by auxiliary classifiers and corresponding true label $y$.
\par
%-----------------------------------------
The logit distillation loss\cite{hinton2015distilling} is also added for student model to learn the output label distribution from teacher as formulated in Eq.~\eqref{logitloss}. By learning the logit information, the student model can make use of it for prediction.
\begin{center}
\begin{equation}
\label{logitloss}
\mathcal{L}_{\text {logit }}=\frac{1}{m}\sum_{i=1}^{m} \sum_{j=1}^{N}L_{K L}\left(c_{j}\left(T_{j}\left(F_{j}^{s}\left(x_{i}\right)\right)\right), c_{j}\left(T_{j}\left(F_{j}^{t}\left(x_{i}\right)\right)\right)\right)
\end{equation}
\end{center}
\par
In Eq.~\eqref{logitloss}, $L_{K L}$ refers to the KL(Kullback-Leibler) divergence, which is used to measure the difference between two probability distributions. It makes the output of student model close to the one of teacher model.
%-------------------loss------------------------
\par
The final loss function in the whole distillation process can be summarized as in Eq.~\eqref{totalloss} with hyper-parameters $\alpha$, $\beta$ and $\gamma$ balancing the proportion of each part:
\begin{center}
\begin{equation}
\label{totalloss}
\mathcal{L}_{\text {total }}= \alpha \cdot \mathcal{L}_{\text {attention }}+\beta \cdot \mathcal{L}_{\text {feature }}+\gamma \cdot \mathcal{L}_{\text {logit }}+\mathcal{L}_{\text {task }}
\end{equation}
\end{center}
\par

\section{Experimental results}
The details of the internal structure and performance of the lightweight network obtained by the designed searching algorithm is presented in this section. The comparison results between the proposed lightweight network and different methods are listed. Furthermore, ablation experiments are conducted to prove the effectiveness of our proposed method. 
\par
\subsection{Dataset and data augmentation}
The dataset used is the public PV cell dataset contributed by the study\cite{ELPV}. There are 2624 EL images of PV cells with resolution of 300$\times$300 pixels, including both monocrystalline and polycrystalline types. 
The images in this public dataset are labeled as defective with a probability. We divide the samples into functional
%functional!
and defective ones with 0.5 as the threshold. And 75\% of the images, i.e. 1970 images are randomly chosen as train set and the left 654 images are test set. All the images are resized to 150$\times$150 pixels. The specific division is shown in Table~\ref{dataset}.
%-----------------dataset table----------------
\begin{table}[ht]
\centering
\renewcommand\arraystretch{1.5} %行高倍数
\scriptsize
\setlength{\tabcolsep}{2mm}{ % 设置宽度
\begin{tabular}{llccr}
\hline
Dataset&Condition&Monocrystalline&Polycrystalline&$\Sigma$\\
\hline
\multirow{2}{*}{Train}&defective&277&340&617\\
&functional&529&824&1353\\
&&&&1970\\
% \hline
\hline
\multirow{2}{*}{Test}&defective&92&112&204\\
&functional&176&274&450\\
&&&&654\\
\hline
\end{tabular}
}
\centering
\caption{Dataset division. The PV cell images are split by a ratio of 75\% for Train set and Test set. The distribution of positive and negative samples and each category of PV cells in the dataset are consistent for Train set and Test set. Notice that data used for NAS process are split from the Train set as the searching train set and the searching test set shown in Figure ~\ref{datasplit}.}
\label{dataset}
\end{table}
\begin{figure}[ht]
    \centering
    \includegraphics[scale=0.75]{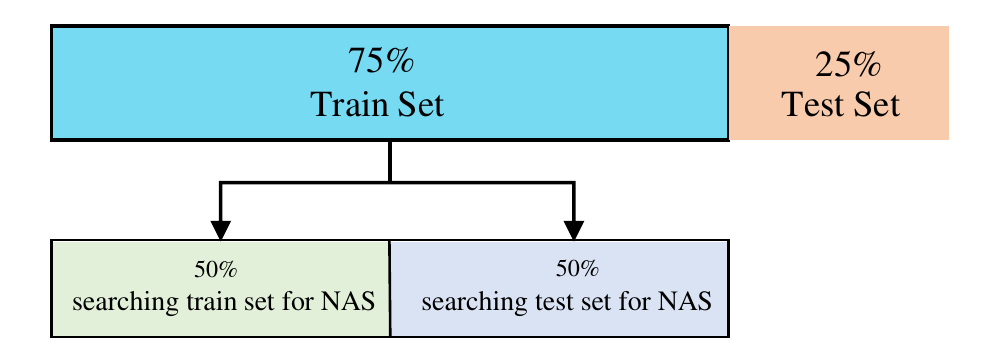}
	\caption{Details of dataset division. The Train set is split into two sets for NAS process with a ratio of 50\%.}
	\label{datasplit} 
\end{figure}
%-------------------------------------------------------
\par
The proportion of positive and negative samples are the same for the train set and the test set. The ratio of the monocrystalline and polycrystalline types are also fixed for train set and test set.
For the lightweight network search process, the whole train set is further divided into searching train set and searching test set for NAS by 50\% each as explained in Figure ~\ref{datasplit}.
\par
Data augmentation is to obtain more representations from the original data without substantially adding data, improving the quality of the original data. It can help the model reduce overfitting and enhance robustness. The data augmentation operations include random horizontal flip, random vertical flip, random rotation within ($-2^{\circ}$,$2^{\circ}$), random rotation within
\{$0^{\circ}$, $90^{\circ}$, $180^{\circ}$, $270^{\circ}$\}
and random affine transformation.
%-------------------------------------------------------------------
% cell fig
\begin{figure*}[ht]
    \centering
    \includegraphics[scale=0.85]{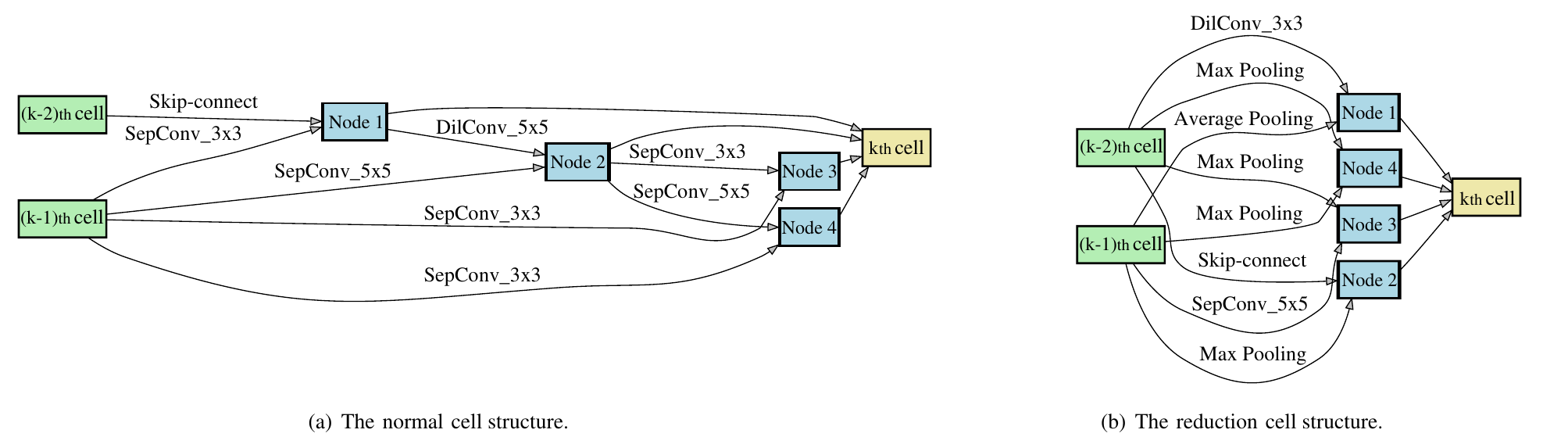}
	\caption{Cell internal structures obtained by searching on public PV cell dataset. The structures are consistent with the schematic diagram in Figure~\ref{cellin}. The detail information of candidate operations can be found in Table~\ref{operations}.}
	\label{cells} 
\end{figure*}
\subsection{Final searched model structure}
The proposed lightweight network is stacked by normal cells and reduction cells.
The final internal architectures of the two kinds of cells by search algorithm are shown in Figure~\ref{cells}.
%------------------------------------------------------------------------------
\subsection{Experimental parameters}
In the process of architecture search, all the convolutional operations follow the order of Rectified linear unit-Convolutional layer-Batch normalization(ReLU-Conv-BN). The cell structure consists of 4 nodes, with two inputs from previous two cells and one output. The initial channels is set as 16. \par
The obtained network by search algorithm tends to choose skip-connect when searching too long, which is called the `Collapse', resulting in poor performance\cite{liang2019darts+}. This is due to the structure variables $\alpha$ and convolution filters parameters $\omega$ in Eq.~\eqref{eq4} competing with each other in the later optimization process. 
Early stopping is an effective way to suppress this phenomenon\cite{liang2019darts+,chen2021pdarts}.
Hence the max search epoch is set as 50 and the number of skip-connect in each cell is limited to less than 2 to avoid deteriorating results.

\par
In knowledge transfer, VGG16(teacher model) and its auxiliary classifiers are trained in advance and are frozen in distillation. The transformation function and auxiliary classifiers chosen for both models are shown in Table~\ref{aux}. The shape of each pair of features is unified through transformation function. The parameter $p$ in Eq.~\eqref{a} is set as 2. The hyper-parameters $\alpha$, $\beta$ and $\gamma$ in Eq.~\eqref{totalloss} are set as 1000, 0.05 and 1 respectly. Our lightweight model is trained in 200 epochs with a batch size of 32. The initial learning rate is set as 0.0025 with weight decay of $7\times10_{}^{-3}$ by stochastic gradient descent(SGD) optimizer.
%------------------------------------------------------------------------
\begin{table}[ht]
\centering
% \begin{threeparttable}
\setlength{\tabcolsep}{3mm}{ % 
\scriptsize
% \rule{0pt}{30pt} %
\renewcommand\arraystretch{1.4} %
\begin{tabular}{l| l |l |c}
\hline
\multirow{2}{*}{Branch} & \multicolumn{2}{c|}{Transformation function}& \multirow{2}{*}{Auxiliary classifier}     \\ \cline{2-3}
&\quad Teacher&\quad Student&\\
\hline
1(shallowest) & SepConv$ \times$3 & SepConv$ \times$3 & \\
2 & SepConv$ \times$2& SepConv$ \times$2 & Pooling\\
3& SepConv & SepConv &Fully Connected Layer\\
4(deepest)&---&Pooling& \\ \hline
% \multirow{4}{*}{Convolutional}
\end{tabular}
}

\caption{Transformation functions and auxiliary classifiers designed for the teacher model and the student model. The serial numbers of branches are sorted by the shallowest to the deepest. Here `SepConv' in transformation functions means depthwise separable convolution.}
\label{aux}
\end{table}
%----------------------------------------
\subsection{Model performance}
In this subsection, we show the performance of the proposed network by quantitative evaluation
 and comparison with the teacher network and other studies. We also evaluate our network by assessment of performance on each category of PV cells. The implementation information is also provided for application reference. To verify the model generalization, we test our model on the private dataset for further demonstrating the effectiveness of the model proposed. 
%--------------------------------------------------------
\subsubsection{Performance comparison}
Our proposed method is compared with 6 manually designed neural networks\cite{ELPV,sun2017defect,bartler2018automated,karimi2019automated,tang2020deep,akram2019cnn} and the teacher model on the public dataset\cite{ELPV} under the same augmentation in 200 epochs. ShuffleNet\cite{ma2018shufflenet} and MobileNet\cite{howard2019mobilenetv3} are also included in experiments as a general efficient light model.
\par
Several traditional feature extraction techniques are also conducted. Open-source algorithms including HOG(histograms of oriented gradients), SIFT(scale-invariant feature transform) and SURF(speeded up robust features) are fed to RBF-kernel SVM classifier for comparison. These algorithms compare the information of centre pixel and neighbourhood pixels to compute a local key point. The parameters of SVM are selected by grid search experiments. 
%------
\begin{table}[t]
\centering
\renewcommand\arraystretch{1} %
\scriptsize
% \tiny
% \resizebox{\textwidth}{5mm}{ % 
\begin{subtable}[h]{1\linewidth}
\setlength{\tabcolsep}{1mm}{ % 
\centering
\begin{tabular}{lcccccr}
\hline
Model&Acc&B\_Acc&Prec&Rec&F1&Parameters\\
&(\%)&(\%)&(\%)&(\%)&(\%)&
\\ \hline
SVM+HOG&85.47&79.65&76.77&66.85&71.47&30.6M
\\
SVM+SIFT&71.85&69.47&23.98&64.37&35.00&0.19M
\\
SVM+SURF&80.03&78.33&24.76&73.71&37.07&0.23M
\\
\hline
Adapted VGG19\cite{ELPV}&87.46&83.92&83.52&74.51&78.76&29.14M\\
Adapted LeNet5\cite{sun2017defect}&80.58&75.04&72.78&60.29&65.95&2.41M\\
Adapted VGG16\cite{bartler2018automated}&82.26&77.87&74.18&66.18&69.95&0.37M\\
CNN\cite{karimi2019automated}&81.35&74.12&78.87&54.90&64.74&0.14M\\
CNN\cite{tang2020deep}&82.56&76.21&79.61&59.31&67.98&12.32M\\
CNN\cite{akram2019cnn}&81.80&76.19&75.76&61.28&67.75&4.73M\\
ShuffleNetV2\cite{ma2018shufflenet}&87.92&85.06&82.72&77.45&80.00&1.24M\\
MobileNetV3\cite{howard2019mobilenetv3}&82.72&75.79&81.82&57.35&67.44&1.22M\\
VGG16(teacher model)&90.52&88.15&86.98&81.86&84.34&32.06M\\
\hline
Ours&\textbf{91.74}&\textbf{90.25}&\textbf{87.13}&\textbf{86.28}&\textbf{86.70}&1.85M\\
\hline
\end{tabular}
\centering
\subcaption{Accuracy, Balanced accuracy, Precision, Recall, F1-score and parameters of ours and other methods.}
}
\end{subtable}
%--------------
\begin{subtable}[h]{1\linewidth}
\setlength{\tabcolsep}{2mm}{ % 
\centering
\begin{tabular}{lcc}
\hline
Model&Acc\_defective&Acc\_functional\\
&(\%)&(\%)
\\ \hline
SVM+HOG&66.85&92.44
\\
SVM+SIFT&63.37&74.58
\\
SVM+SURF&73.71&82.35
\\
\hline
Adapted VGG19\cite{ELPV}&74.51&93.33\\
Adapted LeNet5\cite{sun2017defect}&60.29&89.78\\
Adapted VGG16\cite{bartler2018automated}&66.18&89.56\\
CNN\cite{karimi2019automated}&54.90&93.33\\
CNN\cite{tang2020deep}&59.31&93.11\\
CNN\cite{akram2019cnn}&61.28&91.11\\
ShuffleNetV2\cite{ma2018shufflenet}&77.45&92.67\\
MobileNetV3\cite{howard2019mobilenetv3}&57.35&94.22\\
VGG16(teacher model)&81.86&\textbf{94.44}\\\hline
Ours&\textbf{86.28}&94.22\\
\hline
\end{tabular}
\centering
\subcaption{Accuracy on defective PV cells and functional PV cells respectively of ours and other methods.}
}
\end{subtable}
\centering
\caption{Comparison with other methods on ELPV public dataset under the same data augmentation.}
\label{comparison}
\end{table}
%----------------------------------
\par
These models are modified to classify PV cells into two classes: functional ones or defective ones. 
\par
The quantitative comparisons on performance of the test set and model size are shown in Table~\ref{comparison}. 
In Table~\ref{comparison}.(a), the overall accuracy(Acc) on test set of the proposed model is up to 91.74\%, exceeding other methods. As can be seen, the proposed model achieves or even outperforms the level of teacher model by 1.22\%. It also shows that the proposed model has much less parameters, which can be deployed in practical end devices with less resources than classic large models. 
%-------------
Balanced accuracy(B\_Acc) here means the average of two recall values on defective PV cells and functional PV cells for more fair evaluation. The deep learning based methods tend to perform better than traditional feature extraction on the dataset.
Table~\ref{comparison}.(b) reveals the accuracy of defective PV cells(Acc\_defective) and the accuracy of functional PV cells (Acc\_functional), which reflects performance of recognizing each kinds of PV cells respectively. The ability to correctly recognize defective PV cells is the most core function, which reaches 86.28\% in our network and far exceeds other methods.
\par
%-------------------------------------------------
Comparing with other manually designed models, our proposed network is automatically searched via NAS algorithm with less manual labors. Furthermore, the obtained network achieves the best comprehensive results with a relative light architecture, which proves the effectiveness of the proposed method. 
%------------------
\begin{table}[t]
\centering
\renewcommand\arraystretch{1} %
\scriptsize
\begin{subtable}[h]{1\linewidth}
\setlength{\tabcolsep}{2mm}{ % 
\centering
\begin{tabular}{lccccc}
\multicolumn{6}{l}{\textbf{Performance on Monocrystalline PV Cells}}\\
\hline
Model&Acc&B\_Acc&Prec&Rec&F1\\
&(\%)&(\%)&(\%)&(\%)&(\%)
\\ \hline
SVM+HOG&89.18&86.70&81.82&80.77&81.29\\
SVM+SIFT&73.13&76.52&33.67&84.62&48.21\\
SVM+SURT&82.40&81.45&27.73&79.22&41.08
\\
\hline
Adapted VGG19\cite{ELPV}&86.19&83.78&82.35&76.09&79.10\\
Adapted LeNet5\cite{sun2017defect}&82.09&78.32&78.21&66.30&71.76\\
Adapted VGG16\cite{bartler2018automated}&85.45&82.44&82.72&72.83&77.46\\
CNN\cite{karimi2019automated}&83.58&78.68&85.29&63.04&72.50\\
CNN\cite{tang2020deep}&81.72&74.41&92.16&51.09&65.73\\
CNN\cite{akram2019cnn}&83.96&80.00&82.67&67.39&74.25\\
ShuffleNetV2\cite{ma2018shufflenet}&90.30&88.46&88.37&82.61&85.39\\
MobilenNetV3\cite{howard2019mobilenetv3}&83.96&79.48&84.51&65.22&73.62\\
VGG16(teacher model)&91.42&89.58&90.59&83.70&87.01\\
\hline
Ours&\textbf{93.28}&\textbf{92.30}&\textbf{91.11}&\textbf{89.13}&\textbf{90.11}\\
\hline
\end{tabular}
\centering
\subcaption{Accuracy, Balanced accuracy, Precision, Recall, F1-score and parameters of ours and other methods.}
}
\end{subtable}
%--------------
\begin{subtable}[h]{1\linewidth}
\setlength{\tabcolsep}{2.5mm}{ % 
\centering
\begin{tabular}{lcc}
\multicolumn{3}{l}{\textbf{Performance on Monocrystalline PV Cells}}\\
\hline
Model&Acc\_defective&Acc\_functional\\
&(\%)&(\%)
\\ \hline
SVM+HOG&80.77&92.63\\
SVM+SIFT&84.62&68.42\\
SVM+SIFT&79.22&83.69\\
\hline
Adapted VGG19\cite{ELPV}&76.09&91.48\\
Adapted LeNet5\cite{sun2017defect}&66.30&90.34\\
Adapted VGG16\cite{bartler2018automated}&72.83&92.05\\
CNN\cite{karimi2019automated}&63.04&94.32\\
CNN\cite{tang2020deep}&51.09&97.73\\
CNN\cite{akram2019cnn}&67.39&92.61\\
ShuffleNetV2\cite{ma2018shufflenet}&82.61&94.32\\
MobilenNetV3\cite{howard2019mobilenetv3}&65.22&93.75\\
VGG16(teacher model)&83.70&\textbf{95.46}\\\hline
Ours&\textbf{89.24}&94.22\\
\hline
\end{tabular}
\centering
\subcaption{Accuracy on defective PV cells and functional PV cells respectively of ours and other methods.}
}
\end{subtable}
\centering
\caption{Comparison with other methods on only monocrystalline PV cells of ELPV public dataset under the same data augmentation.}
\label{cmono}
\end{table}
%-----------------------------
% \newpage
\begin{table}[t]
\centering
\renewcommand\arraystretch{1} %
\scriptsize
%--------------
\begin{subtable}[h]{1\linewidth}
\setlength{\tabcolsep}{2mm}{ % 
\centering
\begin{tabular}{lccccc}
\multicolumn{6}{l}{\textbf{Performance on Polycrystalline PV Cells}}\\
\hline
Model&Acc&B\_Acc&Prec&Rec&F1\\
&(\%)&(\%)&(\%)&(\%)&(\%)
\\ \hline
SVM+HOG&82.90&74.15&71.79&56.00&62.92\\
SVM+SIFT&70.94&63.29&16.97&47.92&25.02\\
SVM+SURF&78.91&76.11&22.77&70.41&34.41\\
\hline
Adapted VGG19\cite{ELPV}&88.34&83.87&\textbf{84.54}&73.21&78.47\\
Adapted LeNet5\cite{sun2017defect}&79.53&72.39&68.13&55.36&61.08\\
Adapted VGG16\cite{bartler2018automated}&80.05&74.34&67.33&60.71&63.85\\
CNN\cite{karimi2019automated}&79.79&70.46&72.97&48.21&58.06\\
CNN\cite{tang2020deep}&79.27&73.52&65.69&59.82&62.62\\
CNN\cite{akram2019cnn}&80.31&73.20&70.00&56.25&62.38\\
ShuffleNetV2\cite{ma2018shufflenet}&80.57&80.77&62.76&81.25&70.82\\
MobilenNetV3\cite{howard2019mobilenetv3}&81.87&72.71&79.17&50.89&61.96\\
VGG16(teacher model)&89.90&87.08&84.11&80.36&82.19\\
\hline
Ours&\textbf{90.67}&\textbf{88.68}&83.93&\textbf{83.93}&\textbf{83.93}\\
\hline
\end{tabular}
\centering
\subcaption{Accuracy, Balanced accuracy, Precision, Recall, F1-score and parameters of ours and other methods.}
}
\end{subtable}
%--------------
\begin{subtable}[h]{1\linewidth}
\setlength{\tabcolsep}{2.5mm}{ % 
\centering
\begin{tabular}{lcc}
\multicolumn{3}{l}{\textbf{Performance on Polycrystalline PV Cells}}\\
\hline
Model&Acc\_defective&Acc\_functional\\
&(\%)&(\%)
\\ \hline
SVM+HOG&56.00&92.31\\
SVM+SIFT&47.92&78.67\\
SVM+SURF&70.41&81.82
\\
\hline
Adapted VGG19\cite{ELPV}&73.21&\textbf{94.53}\\
Adapted LeNet5\cite{sun2017defect}&55.36&89.42\\
Adapted VGG16\cite{bartler2018automated}&60.71&87.96\\
CNN\cite{karimi2019automated}&48.21&92.70\\
CNN\cite{tang2020deep}&59.92&87.23\\
CNN\cite{akram2019cnn}&56.25&90.15\\
ShuffleNetV2\cite{ma2018shufflenet}&81.25&80.29\\
MobilenNetV3\cite{howard2019mobilenetv3}&50.89&\textbf{94.53}\\
VGG16(teacher model)&80.36&93.80\\\hline
Ours&\textbf{83.93}&93.43\\
\hline
\end{tabular}
\centering
\subcaption{Accuracy on defective PV cells and functional PV cells respectively of ours and other methods.}
}
\end{subtable}
\centering
\caption{Comparison with other methods on only polycrystalline
PV cells of ELPV public dataset under the same data augmentation.}
\label{cpoly}
\end{table}
%----------------
\subsubsection{Performance on specific categories}
The performance on monocrystalline or polycrystalline PV cells separately of the proposed model are provided to further evaluate the model as shown in Table~\ref{cmono} and Table~\ref{cpoly}. 
\par
On both categories of PV cells, our proposed model has reached the best comprehensive level. On monocrystalline PV cells, every metric of ours achieves the top as described in Table~\ref{cmono}. With regard to the polycrystalline type that is more difficult to deal with, our model can also exceed others by far in Table~\ref{cpoly}, demonstrating the outstanding detection performance on challenging images. Traditional extraction methods tend to perform worse on polycrystalline PV cells, because they focus on high-frequency components of images. It is hard for them to distinguish between cracks and noise of background. 
\subsubsection{Efficiency Comparison}
For end device deployment, a comprehensive consideration needs to be given to the model size and the calculation. To test the efficiency, the proposed model is evaluated on CPU platform(Intel i9-10980XE  24.75M Cache, 3 GHz).
\par
Efficiency comparison, including parameters, FLOPs(floating point operations), inference latency and memory cost of different models are displayed in Table \ref{cpu}. For better comparison, the performances of model balanced accuracy and recall are also included in this table. By the comparison with the second-best network VGG16(the teacher network)
, it can save nearly 180MB memory and 5.6G FLOPs. Traditional methods can run with a fast speed but the performances are poor. Compared with Adapted VGG19 and VGG16(the teacher network), our proposed model requires much less parameters and FLOPs and can reach the comparable speed. At this speed, the proposed model can diagnose 29 samples per second. Even it costs more than other lighter models,  our model is far more accurate than light models\cite{ELPV,sun2017defect,bartler2018automated,karimi2019automated,tang2020deep,akram2019cnn,ma2018shufflenet,howard2019mobilenetv3} by a gap of 3.8\% $\sim$ 11.1\%. 
By the comprehensive consideration of information of resources and performance in Table \ref{cpu}, our proposed model is far superior to other small models and even the large classic networks. The proposed lightweight model can meet the deployment requirements of some common embedded devices with low power consumption, such as Raspberry Pi-4B (4GB, 15W, 9$\sim$10 GFLOPS) and NVIDIA Jetson Nano(4GB, 10W, 7.368 GFLOPS FP64).
%-----------------------------------------
\begin{table}[h]
\centering
\renewcommand\arraystretch{1.2} %
\scriptsize
% \tiny
% \resizebox{\textwidth}{5mm}{ % 
\setlength{\tabcolsep}{1mm}{ % 
\centering
\begin{tabular}{lcccccc}
\hline
Model&B\_Acc&Rec&Parameters&FLOPs&Inference&Memory\\
&(\%)&(\%)&&&Latency&Cost\\
\hline
SVM+HOG&79.65&66.85&30.60M&-&34\textmu s&-\\
SVM+SIFT&69.47&64.37&0.19M&-&74\textmu s&-\\
SVM+SURF&78.33&73.71&0.23M&-&90\textmu s&-
\\
\hline
Adapted VGG19\cite{ELPV}&83.92&74.51&29.14M&8.4G&23ms&194MB\\
Adapted LeNet5\cite{sun2017defect}&75.04&60.29&2.41M&24M&5ms&14MB\\
Adapted VGG16\cite{bartler2018automated}&77.87&66.18&0.37M&108M&5ms&10MB\\
CNN\cite{karimi2019automated}&74.12&54.90&0.14M&50M&2ms&7MB\\
CNN\cite{tang2020deep}&76.21&59.31&12.32M&100M&3ms&54MB\\
CNN\cite{akram2019cnn}&76.19&61.28&4.73M&57M&1ms&22MB\\
ShuffleNetV2\cite{ma2018shufflenet}&85.06&77.45&1.24M&75M&7ms&15MB\\
MobileNetV3\cite{howard2019mobilenetv3}&75.79&57.35&1.22M&32M&6ms&12MB\\
VGG16(teacher model)&88.15&81.86&32.06M&6.7G&29ms&393MB\\\hline
Ours&90.25&86.28&1.85M&1.1G&33ms&214MB\\
\hline
\end{tabular}}
\centering
\caption{Efficiency Comparison on CPU platform(Intel i9-10980XE 24.85M Cache, 3 GHz). FLOPs denotes floating point operations.}
\label{cpu}
\end{table}
%--------------------------------------
\subsubsection{Model generalization ability}
To verify the generalization performance of the models on different data sources, we trained our model on a private PV dataset.
\par
The total 8580 images with 256$\times$256 pixel resolution are extracted from different PV panels of 6$\times$10, 6$\times$12 or 6$\times$24 specifications, containing 482 defective samples and 8098 functional samples. These solar cell samples contain different bus specifications, cell edges and types. Different defects such as material defect, crack, deep crack, disconnected cell interconnect and micro crack are also included in these images. 25\% of images of each class(defective or functional) are randomly selected as test set and the rest as train set.
\par
In terms of the extremely imbalanced class distribution, offline data augmentation is utilized to strengthen learning ability of defect features. Several operations are used on defective ones in train set to avoid overfitting, including horizontal flip, vertical flip, rotation within ($-2^{\circ}$,$2^{\circ}$), rotation within\{$90^{\circ}$, $180^{\circ}$, $270^{\circ}$\}, contrast enhancement, gaussian blur, affine transformation, center cropping, gaussian noise, added black border. These operations tend to simulate actual condition of PV images.
\par
Different methods are trained and tested on this private dataset under the same data augmentation. The results of each model are shown in Table \ref{private}. For the imbalanced distribution of the dataset, balanced accuracy (B\_Acc) and recall of defective PV cells(Acc\_defective) are employed to reflect the model performance precisely. With these extremely imbalanced images from actual PV plants of various sources, our model outperforms teacher model by approximately 2.3\% and 5.7\% in terms of balanced accuracy and accuracy of defective samples, and outperforms other method with a big gap. The accuracy of defective ones of our model is up to 94.26\%, especially showing the better ability in recognizing cell defects in real-world scene.
%-------------------------------------------------------
\begin{table}[h]
\renewcommand\arraystretch{1.2} %
\scriptsize
% \tiny
% \resizebox{\textwidth}{5mm}{ % 
\setlength{\tabcolsep}{1.5mm}{ % 
\begin{tabular}{lccr}
\hline
Model&B\_Acc&Acc\_defective&Parameters\\
&(\%)&(\%)&
\\ \hline
% \textcolor{red}{SVM+HOG}&59.42&63.36&35.08&71.31\\

SVM+HOG&62.91&61.48&46.40M\\
SVM+SIFT&51.52&34.43&0.19M\\
SVM+SURF&53.23&37.70&\textbf{0.10M}
\\
\hline
Adapted VGG19\cite{ELPV}&93.07&86.89&29.14M\\
Adapted LeNet5\cite{sun2017defect}&83.71&71.31&2.41M\\
Adapted VGG16\cite{bartler2018automated}&89.01&79.51&0.37M\\
CNN\cite{karimi2019automated}&76.14&58.20&0.14M\\
CNN\cite{tang2020deep}&77.06&57.38&12.32M\\
CNN\cite{akram2019cnn}&87.87&76.23&4.73M\\
ShuffleNetV2\cite{ma2018shufflenet}&89.77&80.33&1.24M\\
MobileNetV3\cite{howard2019mobilenetv3}&88.58&78.69&1.22M\\
VGG16(teacher model)&92.95&88.53&32.06M\\
\hline
Ours&\textbf{95.23}&\textbf{94.26}&1.85M\\
\hline
\end{tabular}}
\centering
\caption{Comparison with other methods on private dataset under the same data augmentation.}
\label{private}
\end{table}
%-----------------------------------------
\begin{figure*}[ht]
%cam
\centering
\includegraphics[scale=0.55]{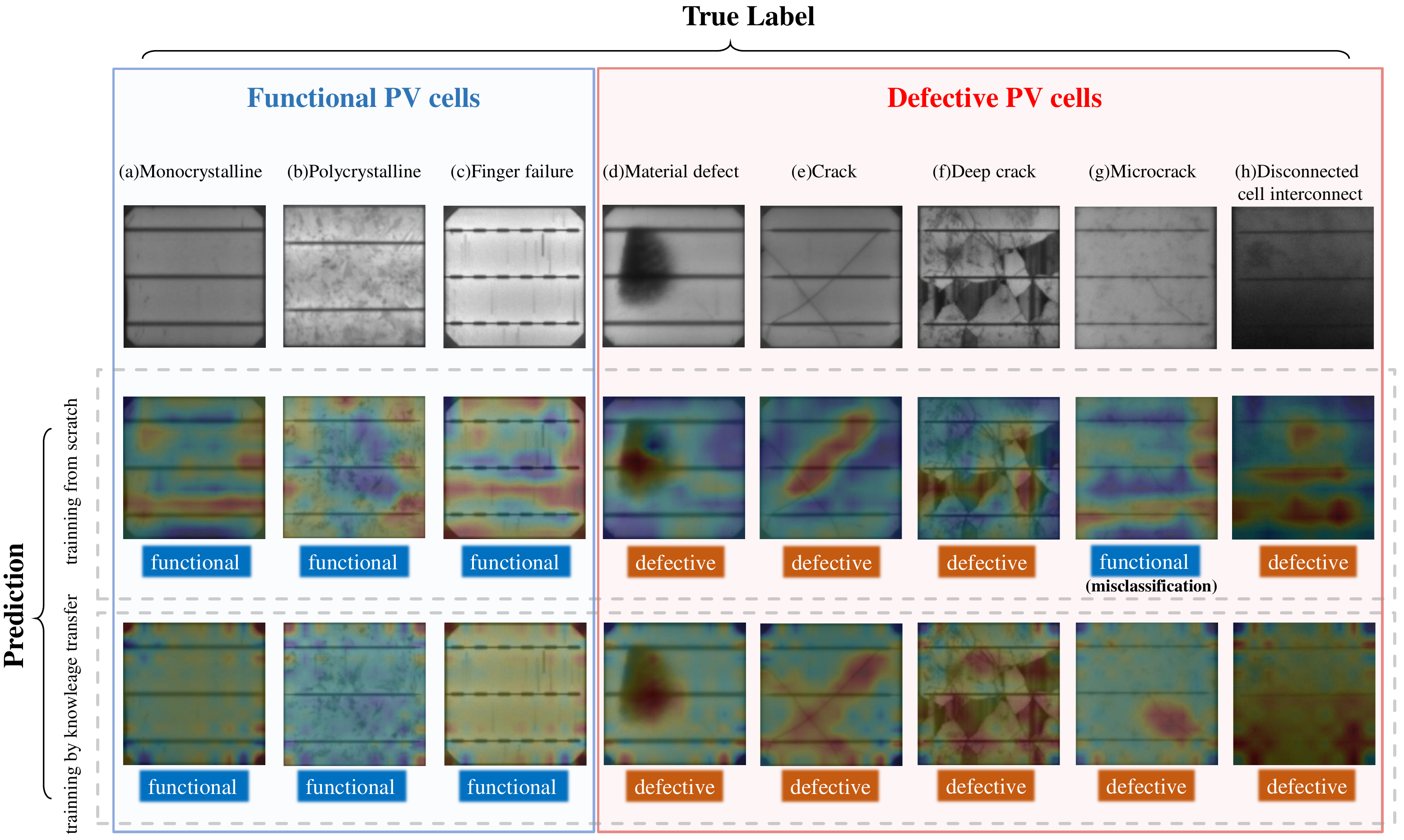}
\caption{The performance on test set of the public PV cell dataset overlaid by Grad-CAM\cite{cam}. Typical images including functional PV cells and defective PV cells are presented. The area that model focuses on is highlighted by heat map.}
\label{images}
\end{figure*}
%--------------------------------
\subsection{Ablation experiments}
In this subsection, we discuss three ablation experiments to demonstrate the effectiveness of our method. Firstly experiments have been carried out to find that training by knowledge transfer achieves better results than training from scratch, demonstrating the importance of prior knowledge in training. Secondly, the roles of different transfer branches are proved in experiments, which verifies the function of both shallow and deep features in transfer process. Finally, the roles of different knowledge priors are illustrated for the function in defect detection.
\subsubsection{The role of prior knowledge}
In our proposed method, the knowledge transfer functions as the prior knowledge which is usually provided by pre-trained large network on large-scale datasets.
\par
In Table~\ref{prior}, the results of models training from scratch are added, which is compared with our proposed one with knowledge transfer. Under the same conditions, there is a big gap between the cases of whether to use the prior knowledge. For F1-score, the performance improves about 13\%, shown in Table~\ref{prior}.(a). The prior knowledge also helps to recognize defective cells, where the accuracy has improved about 22\%, shown in Table~\ref{prior}.(b).
\par
 Figure~\ref{images} shows typical types of samples in the test set of the public dataset\cite{ELPV}, where the columns (a)-(c) show the functional PV cells including finger failure that do not necessarily affect the power efficiency, while the columns (d)-(h) show the typical defects. The original PV cell images are listed in the first row. The second row shows the detection result by the network training from scratch, while the third row displays the results by our proposed method using knowledge transfer. 
 The results using Grad-CAM\cite{cam} show the places that the model pays attention to. The areas by model using knowledge transfer focus more on defects. It shows that the prior knowledge plays a significant role in performance improvement.
\begin{table}[h]
\centering
\renewcommand\arraystretch{1.2} %
\scriptsize
% \tiny
% \resizebox{\textwidth}{5mm}{ % 
% t
% \begin{subtable}[t]{0.8\linewidth}
\begin{subtable}[g]{1\linewidth}
\setlength{\tabcolsep}{2mm}{ % 
\centering
\begin{tabular}{lccccc}
\hline
 with Prior&Acc&B\_Acc&Prec&Rec&F1\\
Knowledge&(\%)&(\%)&(\%)&(\%)&(\%)
\\ \hline
\quad$\times$&85.63&79.77&86.18&64.22&73.60\\
\quad$\checkmark$(ours)&\textbf{91.74}&\textbf{90.25}&\textbf{87.13}&\textbf{86.28}&\textbf{86.70}\\
\hline
\end{tabular}
\centering
\subcaption{Accuracy, Balanced accuracy, Precision, Recall, F1-score and parameters of the proposed model training in different ways.}
}
\end{subtable}
\begin{subtable}[g]{1\linewidth}
\setlength{\tabcolsep}{1.6mm}{ % 
\centering
\begin{tabular}{lcc}
\hline
 with Prior &Acc\_defective&Acc\_functional\\
Knowledge&(\%)&(\%)
\\ \hline
\quad$\times$&64.22&\textbf{95.33}\\
\quad$\checkmark$(ours)&\textbf{86.28}&94.22\\
\hline
\end{tabular}
\subcaption{Accuracy on defective PV cells and functional PV cells respectively of the proposed model training in different ways.}
}
\end{subtable}
\centering
\caption{Ablation study of using prior knowledge in model training. Training from scratch and training using knowledge transfer are denoted by `$\times$' and`$\checkmark$' respectively.}
\label{prior}
\end{table}
%--------------------------------------------------

\begin{table}[t]
\centering
\renewcommand\arraystretch{1.2} %
\scriptsize
\begin{subtable}[g]{1\linewidth}
\setlength{\tabcolsep}{1.5mm}{ % 
\centering
\begin{tabular}{ccclccccc}
\hline
\multicolumn{4}{c}{Branch}&Acc&B\_Acc&Prec&Rec&F1\\
1(shallowest)&2&3&4(deepest)&(\%)&(\%)&(\%)&(\%)&(\%) 
\\
\hline
\quad$\times$&\checkmark\quad&\checkmark&\quad\checkmark&89.91&87.58&85.57&81.37&83.42\\
\quad\checkmark&$\times$\quad&\checkmark&\quad\checkmark&89.60&87.49&84.34&81.86&83.08\\
\quad\checkmark&\checkmark\quad&$\times$&\quad\checkmark&88.99&86.91&83.00&81.37&82.18\\
\quad\checkmark&\checkmark\quad&\checkmark&\quad$\times$&88.84&86.53&83.25&80.39&81.80\\
\hline
\quad\checkmark&\checkmark\quad&\checkmark&\quad\checkmark 
\quad(ours)&\textbf{91.74}&\textbf{90.25}&\textbf{87.13}&\textbf{86.28}&\textbf{86.70}\\
\hline
\end{tabular}
\subcaption{Accuracy, Balanced accuracy, Precision, Recall, F1-score and parameters of the proposed model using features at different depths.}
}
\end{subtable}
\begin{subtable}[g]{1\linewidth}
\setlength{\tabcolsep}{2mm}{ 
\centering
\begin{tabular}{ccclcc}
\hline
\multicolumn{4}{c}{Branch}&Acc\_defective&Acc\_functional\\
1(shallowest)&2&3&4(deepest)&(\%)&(\%)
\\
\hline
$\times$&\checkmark&\checkmark&\quad\checkmark&81.37&93.78\\

\checkmark&$\times$&\checkmark&\quad\checkmark&81.86&93.11\\

\checkmark&\checkmark&$\times$&\quad\checkmark&81.37&92.44\\

\checkmark&\checkmark&\checkmark&\quad$\times$&80.39&92.67\\
% $\surd$&$\surd$&$\surd$&$\times$&88.84&80.39&92.67&86.53&83.25&80.39&81.80\\
\hline
\checkmark&\checkmark&\checkmark&\quad\checkmark\quad (ours)&\textbf{86.28}&\textbf{94.22}\\
\hline
\end{tabular}
\subcaption{Accuracy on defective PV cells and functional PV cells respectively of the proposed model using features at different depths.}
}
\end{subtable}

\centering
\caption{Ablation study of branches at different depths in knowledge transfer. Transfer branches used are denoted by `$\checkmark$' and removed ones are marked by `$\times$'.}
\label{depth_ab}
\end{table}
%------
\subsubsection{The role of knowledge at different depths in knowledge transfer}
Feature maps with different resolution represent various information including deep semantics and shallow details of the objects. To grasp abundant features of the PV cells, the auxiliary classifiers are attached at different depths. Table~\ref{depth_ab} shows the role of each auxiliary classifier, where the serial numbers of branches are sorted by the shallowest to the deepest.
As is presented, the deeper branch plays a more important role in defect detection. The lack of any branch can lead to degradation of the model, which proves the function of features at different depths in knowledge transfer.

%---------------------------------
\begin{table}[t]
\centering
\renewcommand\arraystretch{1.2} %
\scriptsize
% \tiny
% \resizebox{\textwidth}{5mm}{ % 
% t
\begin{subtable}[t]{1\linewidth}
\setlength{\tabcolsep}{1.2mm}{ % 
\centering
\begin{tabular}{ccclccccc}
\hline
\multicolumn{4}{c}{$\mathcal{L}_{\text {total }}\qquad$}&Acc&B\_Acc&Prec&Rec&F1\\
% $L_{2}$&$L_{AT}$
$\mathcal{L}_{attention}$&$\mathcal{L}_{feature}$&$\mathcal{L}_{logit}$&$\mathcal{L}_{task}$&(\%)&(\%)&(\%)&(\%)&(\%) 
\\
\hline
$\times$&\checkmark&\checkmark&\quad\checkmark&89.91&88.51&83.17&84.80&83.98\\

\checkmark&$\times$&\checkmark&\quad\checkmark&90.37&87.64&87.70&80.39&83.89\\

\checkmark&\checkmark&$\times$&\quad\checkmark&89.30&87.67&82.52&83.33&82.93\\

\checkmark&\checkmark&\checkmark&\quad$\times$&\textbf{92.20}&89.24&\textbf{92.74}&81.37&86.68\\

\hline
\checkmark&\checkmark&\checkmark&\quad\checkmark\quad(ours)&91.74&\textbf{90.25}&87.13&\textbf{86.28}&\textbf{86.70}\\
\hline
\end{tabular}
\subcaption{Accuracy, Balanced accuracy, Precision, Recall, F1-score and parameters of the proposed model using different knowledge.}
}
\end{subtable}
%--------
\begin{subtable}[t]{1\linewidth}
\setlength{\tabcolsep}{1.8mm}{ % 设置宽度
\centering
\begin{tabular}{ccclcc}
\hline
\multicolumn{4}{c}{$\mathcal{L}_{\text {total }}\qquad$}&Acc\_defective&Acc\_functional\\

$\mathcal{L}_{attention}$&$\mathcal{L}_{feature}$&$\mathcal{L}_{logit}$&$\mathcal{L}_{task}$&(\%)&(\%)
\\
\hline
% no at
$\times$&\checkmark&\checkmark&\checkmark&84.80&92.22\\
%no l2
\checkmark&$\times$&\checkmark&\checkmark&80.39&94.89\\
%no logit
\checkmark&\checkmark&$\times$&\checkmark&83.33&92.00\\
% no task
\checkmark&\checkmark&\checkmark&$\times$&81.37&\textbf{97.11}\\

\hline
\checkmark&\checkmark&\checkmark&\checkmark\quad(ours)&\textbf{86.28}&94.22\\
\hline
\end{tabular}
\subcaption{Accuracy on defective PV cells and functional PV cells respectively of the proposed model using different knowledge.}
}
\end{subtable}

\centering
\caption{Ablation study of different knowledge in knowledge transfer. Knowledge used in training are denoted by `$\checkmark$' and ones marked by `$\times$' are not added in loss function.}
\label{loss_ab}
\end{table}
%---------------------------
\subsubsection{The role of different knowledge in knowledge transfer}
We exploit different kinds of prior knowledges to improve the performance of the lightweight model on task of defective PV cell detection, including attention information, feature information, logit information and task-oriented information. The roles of different knowledge are evaluated, as shown in Table~\ref{loss_ab}. It shows that the task loss takes effect in the rise of the recall at the expense of some other performance, effectively increasing the ability of recognize the defects on PV cells, which is significant to the practical application. Adding attention knowledge also brings the increase of the performance.
\par
The model combined all kinds of loss performs top on most metrics especially the recall.

\section{Conclusion}
%-------------------------------------------
\par
We have proposed a novel approach to acquire a lightweight network for detection of defective PV cells using EL images.
The proposed network achieves the state-of-the-art performance on the public PV cell dataset of EL images under online data augmentation with high accuracy of 91.74\% and 1.85M parameters. It requires less computation and hardware sources, and retains superior performance at the same time. Especially the ability on recognizing defective PV cells appears much higher.
\par
 To automatically obtain the lightweight network, we exploit the NAS algorithm to search the network with less manual workload. Based on the multi-scale characteristics of PV cells, the search space is designed to utilize useful scale information. Furthermore, training the proposed lightweight model from scratch can not well utilize useful prior knowledge. To make full use of the priors already learned by the large-scale network, knowledge distillation is utilized and various kinds of knowledge are considered into the transfer process. The ablation experiments are also conducted to prove the effectiveness of our method. Experiments and evaluation on both public dataset and private dataset have demonstrated that our model exceeds other method by much better performance and relatively lighter size, which shows that it can be an effective tool for practical applications and terminal deployment in field and industry. The proposed method has provided a new idea of designing models for application scenes.
 \par
 In future, different types of defects can be considered into further study. The way of obtaining lightweight models on segmentation task can be further considered, because specific location or segmentation result of defects can provide more benefit for practical deployment in industry. Existing segmentation models can be hard to be applied in PV cell defect detection task for their larger network volume. Computing acceleration on terminals will also be investigated for further network compression and speeding up.
\label{}

\bibliographystyle{elsarticle-num} 
\bibliography{ref}

\begin{thebibliography}{10}
\expandafter\ifx\csname url\endcsname\relax
  \def\url#1{\texttt{#1}}\fi
\expandafter\ifx\csname urlprefix\endcsname\relax\def\urlprefix{URL }\fi
\expandafter\ifx\csname href\endcsname\relax
  \def\href#1#2{#2} \def\path#1{#1}\fi

\bibitem{NDIAYE2013140}
A.~Ndiaye, A.~Charki, A.~Kobi, C.~M. Kébé, P.~A. Ndiaye, V.~Sambou,
  Degradations of silicon photovoltaic modules: A literature review, Solar
  Energy 96 (2013) 140--151.
\newblock \href {https://doi.org/10.1016/j.solener.2013.07.005}
  {\path{doi:10.1016/j.solener.2013.07.005}}.

\bibitem{li2019thermo}
G.~Li, M.~Akram, Y.~Jin, X.~Chen, C.~Zhu, A.~Ahmad, R.~Arshad, X.~Zhao,
  Thermo-mechanical behavior assessment of smart wire connected and busbarpv
  modules during production, transportation, and subsequent field loading
  stages, Energy 168 (2019) 931--945.
\newblock \href {https://doi.org/10.1016/j.energy.2018.12.002}
  {\path{doi:10.1016/j.energy.2018.12.002}}.

\bibitem{kontges2014review}
M.~K{\"o}ntges, S.~Kurtz, C.~Packard, U.~Jahn, K.~Berger, K.~Kato, T.~Friesen,
  H.~Liu, M.~Van~Iseghem, J.~Wohlgemuth, et~al.,
  \href{https://repository.supsi.ch/9645/}{Review of Failures of Photovoltaic
  Modules}, IEA-Photovoltaic Power Systems Programme Technical Report 13-01
  (2014) 1--140.

\bibitem{kajari2012criticality}
S.~Kajari-Schr{\v{s}}der, I.~Kunze, M.~K{\v{s}}ntges, Criticality of cracks in
  pv modules, Energy Procedia 27 (2012) 658--663.
\newblock \href {https://doi.org/10.1016/j.egypro.2012.07.125}
  {\path{doi:10.1016/j.egypro.2012.07.125}}.

\bibitem{dhimish2020micro}
M.~Dhimish, Micro cracks distribution and power degradation of polycrystalline
  solar cells wafer: Observations constructed from the analysis of 4000
  samples, Renewable Energy 145 (2020) 466--477.
\newblock \href {https://doi.org/10.1016/j.renene.2019.06.057}
  {\path{doi:10.1016/j.renene.2019.06.057}}.

\bibitem{abdelhamid2013review}
M.~Abdelhamid, R.~Singh, M.~Omar, Review of microcrack detection techniques for
  silicon solar cells, IEEE Journal of Photovoltaics 4~(1) (2013) 514--524.
\newblock \href {https://doi.org/10.1109/JPHOTOV.2013.2285622}
  {\path{doi:10.1109/JPHOTOV.2013.2285622}}.

\bibitem{TSANAKAS2016695}
J.~A. Tsanakas, L.~Ha, C.~Buerhop, Faults and infrared thermographic diagnosis
  in operating c-si photovoltaic modules: A review of research and future
  challenges, Renewable and Sustainable Energy Reviews 62 (2016) 695--709.
\newblock \href {https://doi.org/10.1016/j.rser.2016.04.079}
  {\path{doi:10.1016/j.rser.2016.04.079}}.

\bibitem{fuyuki2009photographic}
T.~Fuyuki, A.~Kitiyanan, Photographic diagnosis of crystalline silicon solar
  cells utilizing electroluminescence, Applied Physics A 96~(1) (2009)
  189--196.
\newblock \href {https://doi.org/10.1007/s00339-008-4986-0}
  {\path{doi:10.1007/s00339-008-4986-0}}.

\bibitem{breitenstein2011can}
O.~Breitenstein, J.~Bauer, K.~Bothe, D.~Hinken, J.~M{\"u}ller, W.~Kwapil, M.~C.
  Schubert, W.~Warta, Can luminescence imaging replace lock-in thermography on
  solar cells, IEEE Journal of Photovoltaics 1~(2) (2011) 159--167.
\newblock \href {https://doi.org/10.1109/JPHOTOV.2011.2169394}
  {\path{doi:10.1109/JPHOTOV.2011.2169394}}.

\bibitem{karimi2019automated}
A.~M. Karimi, J.~S. Fada, M.~A. Hossain, S.~Yang, T.~J. Peshek, J.~L. Braid,
  R.~H. French, Automated pipeline for photovoltaic module electroluminescence
  image processing and degradation feature classification, IEEE Journal of
  Photovoltaics 9~(5) (2019) 1324--1335.
\newblock \href {https://doi.org/10.1109/JPHOTOV.2019.2920732}
  {\path{doi:10.1109/JPHOTOV.2019.2920732}}.

\bibitem{tang2020deep}
W.~Tang, Q.~Yang, K.~Xiong, W.~Yan, Deep learning based automatic defect
  identification of photovoltaic module using electroluminescence images, Solar
  Energy 201 (2020) 453--460.
\newblock \href {https://doi.org/10.1016/j.solener.2020.03.049}
  {\path{doi:10.1016/j.solener.2020.03.049}}.

\bibitem{akram2019cnn}
M.~W. Akram, G.~Li, Y.~Jin, X.~Chen, C.~Zhu, X.~Zhao, A.~Khaliq, M.~Faheem,
  A.~Ahmad, Cnn based automatic detection of photovoltaic cell defects in
  electroluminescence images, Energy 189 (2019) 116319.
\newblock \href {https://doi.org/10.1016/j.energy.2019.116319}
  {\path{doi:10.1016/j.energy.2019.116319}}.

\bibitem{ELPV}
S.~Deitsch, V.~Christlein, S.~Berger, C.~Buerhop-Lutz, A.~Maier, F.~Gallwitz,
  C.~Riess, Automatic classification of defective photovoltaic module cells in
  electroluminescence images, Solar Energy 185 (2019) 455--468.
\newblock \href {https://doi.org/10.1016/j.solener.2019.02.067}
  {\path{doi:10.1016/j.solener.2019.02.067}}.

\bibitem{dhimish2019novel}
M.~Dhimish, V.~Holmes, P.~Mather, Novel photovoltaic micro crack detection
  technique, IEEE Transactions on Device and Materials Reliability 19~(2)
  (2019) 304--312.
\newblock \href {https://doi.org/10.1109/TDMR.2019.2907019}
  {\path{doi:10.1109/TDMR.2019.2907019}}.

\bibitem{dhimish2019solar}
M.~Dhimish, V.~Holmes, Solar cells micro crack detection technique using
  state-of-the-art electroluminescence imaging, Journal of Science: Advanced
  Materials and Devices 4~(4) (2019) 499--508.
\newblock \href {https://doi.org/10.1016/j.jsamd.2019.10.004}
  {\path{doi:10.1016/j.jsamd.2019.10.004}}.

\bibitem{tsai2012defect}
D.-M. Tsai, S.-C. Wu, W.-Y. Chiu, Defect detection in solar modules using ica
  basis images, IEEE Transactions on Industrial Informatics 9~(1) (2012)
  122--131.
\newblock \href {https://doi.org/10.1109/TII.2012.2209663}
  {\path{doi:10.1109/TII.2012.2209663}}.

\bibitem{anwar2014micro}
S.~A. Anwar, M.~Z. Abdullah, Micro-crack detection of multicrystalline solar
  cells featuring an improved anisotropic diffusion filter and image
  segmentation technique, EURASIP Journal on Image and Video Processing
  2014~(1) (2014) 1--17.
\newblock \href {https://doi.org/10.1186/1687-5281-2014-15}
  {\path{doi:10.1186/1687-5281-2014-15}}.

\bibitem{su2019classification}
B.~Su, H.~Chen, Y.~Zhu, W.~Liu, K.~Liu, Classification of manufacturing defects
  in multicrystalline solar cells with novel feature descriptor, IEEE
  Transactions on Instrumentation and Measurement 68~(12) (2019) 4675--4688.
\newblock \href {https://doi.org/10.1109/TIM.2019.2900961}
  {\path{doi:10.1109/TIM.2019.2900961}}.

\bibitem{sun2017defect}
M.~Sun, S.~Lv, X.~Zhao, R.~Li, W.~Zhang, X.~Zhang, Defect detection of
  photovoltaic modules based on convolutional neural network, in: International
  Conference on Machine Learning and Intelligent Communications, Springer,
  2017, pp. 122--132.
\newblock \href {https://doi.org/10.1007/978-3-319-73564-1_13}
  {\path{doi:10.1007/978-3-319-73564-1_13}}.

\bibitem{LeNet5}
Y.~Lecun, L.~Bottou, Y.~Bengio, P.~Haffner, Gradient-based learning applied to
  document recognition, Proceedings of the IEEE 86~(11) (1998) 2278--2324.
\newblock \href {https://doi.org/10.1109/5.726791}
  {\path{doi:10.1109/5.726791}}.

\bibitem{bartler2018automated}
A.~Bartler, L.~Mauch, B.~Yang, M.~Reuter, L.~Stoicescu, Automated detection of
  solar cell defects with deep learning, in: 2018 26th European signal
  processing conference (EUSIPCO), IEEE, 2018, pp. 2035--2039.
\newblock \href {https://doi.org/10.23919/EUSIPCO.2018.8553025}
  {\path{doi:10.23919/EUSIPCO.2018.8553025}}.

\bibitem{vgg}
K.~Simonyan, A.~Zisserman, Very deep convolutional networks for large-scale
  image recognition, arXiv preprint arXiv:1409.1556 (2014).
\newblock \href {http://arxiv.org/abs/1409.1556} {\path{arXiv:1409.1556}}.

\bibitem{deitsch2019automatic}
S.~Deitsch, V.~Christlein, S.~Berger, C.~Buerhop-Lutz, A.~Maier, F.~Gallwitz,
  C.~Riess, Automatic classification of defective photovoltaic module cells in
  electroluminescence images, Solar Energy 185 (2019) 455--468.
\newblock \href {https://doi.org/10.1016/j.solener.2019.02.067}
  {\path{doi:10.1016/j.solener.2019.02.067}}.

\bibitem{shou2020defect}
C.~Shou, L.~Hong, W.~Ding, Q.~Shen, W.~Zhou, Y.~Jiang, C.~Zhao, Defect
  detection with generative adversarial networks for electroluminescence images
  of solar cells, in: 2020 35th Youth Academic Annual Conference of Chinese
  Association of Automation (YAC), IEEE, 2020, pp. 312--317.
\newblock \href {https://doi.org/10.1109/YAC51587.2020.9337676}
  {\path{doi:10.1109/YAC51587.2020.9337676}}.

\bibitem{liu2019surface}
L.~Liu, Y.~Zhu, M.~R.~U. Rahman, P.~Zhao, H.~Chen, Surface defect detection of
  solar cells based on feature pyramid network and ga-faster-rcnn, in: 2019 2nd
  China Symposium on Cognitive Computing and Hybrid Intelligence (CCHI), IEEE,
  2019, pp. 292--297.
\newblock \href {https://doi.org/10.1109/CCHI.2019.8901952}
  {\path{doi:10.1109/CCHI.2019.8901952}}.

\bibitem{su2020deep}
B.~Su, H.~Chen, P.~Chen, G.~Bian, K.~Liu, W.~Liu, Deep learning-based
  solar-cell manufacturing defect detection with complementary attention
  network, IEEE Transactions on Industrial Informatics 17~(6) (2020)
  4084--4095.
\newblock \href {https://doi.org/10.1109/TII.2020.3008021}
  {\path{doi:10.1109/TII.2020.3008021}}.

\bibitem{WANG2022119203}
H.~Wang, H.~Chen, B.~Wang, Y.~Jin, G.~Li, Y.~Kan, High-efficiency low-power
  microdefect detection in photovoltaic cells via a field programmable gate
  array-accelerated dual-flow network, Applied Energy 318 (2022) 119203.
\newblock \href {https://doi.org/10.1016/j.apenergy.2022.119203}
  {\path{doi:10.1016/j.apenergy.2022.119203}}.

\bibitem{liu2018darts}
H.~Liu, K.~Simonyan, Y.~Yang, Darts: Differentiable architecture search, arXiv
  preprint arXiv:1806.09055 (2018).
\newblock \href {http://arxiv.org/abs/1806.09055} {\path{arXiv:1806.09055}}.

\bibitem{resnet}
K.~He, X.~Zhang, S.~Ren, J.~Sun, Deep residual learning for image recognition,
  in: Proceedings of the IEEE conference on computer vision and pattern
  recognition, 2016, pp. 770--778.
\newblock \href {https://doi.org/10.1109/CVPR.2016.90}
  {\path{doi:10.1109/CVPR.2016.90}}.

\bibitem{hinton2015distilling}
G.~Hinton, O.~Vinyals, J.~Dean, et~al., Distilling the knowledge in a neural
  network, arXiv preprint arXiv:1503.02531 2~(7) (2015).
\newblock \href {http://arxiv.org/abs/1503.02531} {\path{arXiv:1503.02531}}.

\bibitem{gotmare2018closer}
A.~Gotmare, N.~S. Keskar, C.~Xiong, R.~Socher, A closer look at deep learning
  heuristics: Learning rate restarts, warmup and distillation, arXiv preprint
  arXiv:1810.13243 (2018).
\newblock \href {http://arxiv.org/abs/1810.13243} {\path{arXiv:1810.13243}}.

\bibitem{romero2014fitnets}
A.~Romero, N.~Ballas, S.~E. Kahou, A.~Chassang, C.~Gatta, Y.~Bengio, Fitnets:
  Hints for thin deep nets, arXiv preprint arXiv:1412.6550 (2014).
\newblock \href {http://arxiv.org/abs/1412.6550} {\path{arXiv:1412.6550}}.

\bibitem{zhang2020TOFD}
L.~Zhang, Y.~Shi, Z.~Shi, K.~Ma, C.~Bao,
  \href{https://proceedings.neurips.cc/paper/2020/file/a96b65a721e561e1e3de768ac819ffbb-Paper.pdf}{Task-oriented
  feature distillation}, Advances in Neural Information Processing Systems 33
  (2020) 14759--14771.

\bibitem{zagoruyko2016AT}
S.~Zagoruyko, N.~Komodakis, Paying more attention to attention: Improving the
  performance of convolutional neural networks via attention transfer, arXiv
  preprint arXiv:1612.03928 (2016).
\newblock \href {http://arxiv.org/abs/1612.03928} {\path{arXiv:1612.03928}}.

\bibitem{liang2019darts+}
H.~Liang, S.~Zhang, J.~Sun, X.~He, W.~Huang, K.~Zhuang, Z.~Li, Darts+: Improved
  differentiable architecture search with early stopping, arXiv preprint
  arXiv:1909.06035 (2019).
\newblock \href {http://arxiv.org/abs/1909.06035} {\path{arXiv:1909.06035}}.

\bibitem{chen2021pdarts}
X.~Chen, L.~Xie, J.~Wu, Q.~Tian, Progressive darts: Bridging the optimization
  gap for nas in the wild, International Journal of Computer Vision 129~(3)
  (2021) 638--655.
\newblock \href {https://doi.org/10.1007/s11263-020-01396-x}
  {\path{doi:10.1007/s11263-020-01396-x}}.

\bibitem{ma2018shufflenet}
N.~Ma, X.~Zhang, H.-T. Zheng, J.~Sun,
  \href{https://openaccess.thecvf.com/content_ECCV_2018/papers/Ningning_Light-weight_CNN_Architecture_ECCV_2018_paper.pdf}{Shufflenet
  V2: Practical guidelines for efficient CNN architecture design}, in:
  Proceedings of the European conference on computer vision (ECCV), 2018, pp.
  116--131.

\bibitem{howard2019mobilenetv3}
A.~Howard, M.~Sandler, G.~Chu, L.-C. Chen, B.~Chen, M.~Tan, W.~Wang, Y.~Zhu,
  R.~Pang, V.~Vasudevan, et~al.,
  \href{https://openaccess.thecvf.com/content_ICCV_2019/papers/Howard_Searching_for_MobileNetV3_ICCV_2019_paper.pdf}{Searching
  for MobileNetV3}, in: Proceedings of the IEEE/CVF international conference on
  computer vision, 2019, pp. 1314--1324.

\bibitem{cam}
R.~R. Selvaraju, A.~Das, R.~Vedantam, M.~Cogswell, D.~Parikh, D.~Batra,
  Grad-cam: Why did you say that? visual explanations from deep networks via
  gradient-based localization, arXiv preprint arXiv:1610.02391 (2016).
\newblock \href {http://arxiv.org/abs/1610.02391} {\path{arXiv:1610.02391}}.

\end{thebibliography}
% \end{thebibliography}
\end{document}